\documentclass[runningheads]{llncs}

\usepackage{eccv}

\usepackage{eccvabbrv}

\usepackage{graphicx}
\usepackage{booktabs}    
\usepackage{multirow}    
\usepackage{amssymb}     
\usepackage[table]{xcolor} 
\usepackage{tabularx}    
\usepackage{microtype}   
\usepackage{subcaption}  
\usepackage{pifont}
\usepackage{makecell}
\usepackage{marvosym}   
\usepackage[accsupp]{axessibility}

\usepackage{hyperref}
\usepackage{bookmark}
\usepackage{orcidlink}

\begin{document}

\title{CritiqueDriveVLM: From Verifier-Guided Reinforcement Learning to Latent Thought Distillation for Autonomous Driving} 

\titlerunning{CritiqueDriveVLM}




\author{Zhaohong Liu\inst{1,3} \and
Hao Ye\inst{1,3} \and
Xianlin Zhang\inst{2,3} \and
Mengshi Qi\inst{1,3}\textsuperscript{(\Letter)}}

\authorrunning{Z.~Liu et al.}

\institute{State Key Laboratory of Networking and Switching Technology
\and
School of Digital Media \& Design Arts
\and
Beijing University of Posts and Telecommunications, Beijing, China \\
\email{\{liuzhaoh, haoye, zxlin, qms\}@bupt.edu.cn}}

\maketitle

\begin{abstract}
End-to-end Vision-Language Models (VLMs) show immense potential in autonomous driving. However, standard Supervised Fine-Tuning (SFT) often suffers from reasoning hallucinations and conservative biases. While traditional tool-augmented frameworks and Chain-of-Thought (CoT) approaches mitigate these issues, they incur exorbitant token consumption and unacceptable latency, rendering real-time deployment impractical. To resolve this reliability-efficiency trade-off, we propose \textbf{CritiqueDriveVLM}, a novel unified three-stage framework internalizing reasoning directly into the VLM. First, we introduce Critique-Driven Multi-Turn Reinforcement Learning (RL) guided by a multi-dimensional verifier. By providing granular scalar feedback and a multi-turn penalty, we force the policy to internalize logical deduction, cultivating a robust System-2 Teacher that achieves high accuracy without fragile external tools. Subsequently, we propose Latent Thought Distillation to overcome the latency bottleneck. By aligning the Student's latent representations with the Teacher's fully converged reasoning states, we compress deep logical capabilities into a fast, CoT-free System-1 Student. Extensive experiments on the widely-used DriveLMM-o1 benchmark demonstrate remarkable improvements. Compared to the base model, our tool-free Teacher significantly boosts Multiple Choice Quality (MCQ) from 55.54\% to a state-of-the-art 76.54\%. Crucially, our distilled Student preserves competitive reasoning depth while drastically minimizing generation length to an average of merely 28 tokens. This slashes inference latency by 88\% (from 3482 ms to 416 ms), paving a highly robust pathway for low-latency autonomous driving. Our source code is available at \url{https://github.com/MICLAB-BUPT/CritiqueDriveVLM}.
  
\end{abstract}

\begin{figure}[tb]
  \centering
  \includegraphics[width=0.98\textwidth]{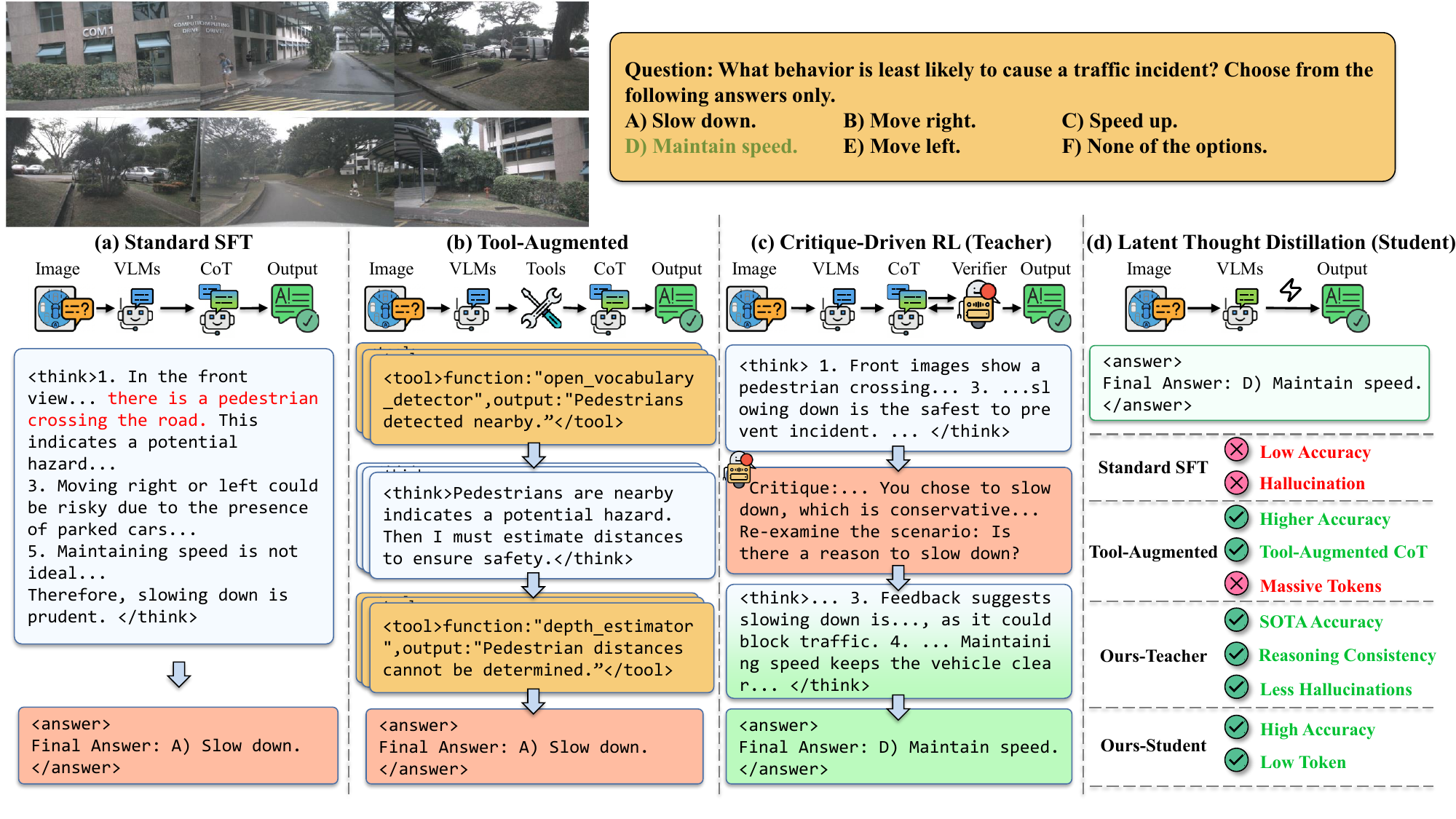}
  \caption{
Paradigm comparison of VLM-based autonomous driving. (a) Standard SFT is prone to reasoning hallucinations and conservative biases. (b) Tool-Augmented methods suffer from brittle external APIs and high latency. (c) Critique-Driven RL (Teacher) internalizes deep logic without relying on external tools. (d) Latent Thought Distillation (Student) enables instant, CoT-free execution, eliminating the overhead of explicit reasoning tokens.
}
  \label{fig:intro}
\end{figure}

\section{Introduction}
\label{sec:intro}

The integration of Vision-Language Models (VLMs)~\cite{liu2023llava,radford2021learning,bai2025qwen3,chen2025internvl,bai2025qwen25vl} has catalyzed a profound paradigm shift in autonomous driving, transitioning the field from traditional, modular perception-planning pipelines~\cite{fan2018baidu,levinson2011towards,li2022bevformer,gao2020vectornet} toward end-to-end holistic reasoning frameworks~\cite{jiang2023vad,hu2023planning}. By leveraging extensive world knowledge and powerful visual-semantic alignment, VLMs demonstrate remarkable potential in interpreting complex traffic scenes and understanding nuanced dynamic interactions. However, deploying these models directly into safety-critical environments remains an open challenge. Current methodologies predominantly rely on standard Supervised Fine-Tuning (SFT)~\cite{shao2024lmdrive} to align pre-trained VLMs with driving-specific instructions. Unfortunately, standard SFT merely mimics the surface-level statistical distribution of human driving trajectories without genuinely internalizing deep logical deduction capabilities. Consequently, in complex dynamic scenarios, SFT models frequently suffer from severe reasoning hallucinations and conservative biases. As illustrated in~\Cref{fig:intro}(a), a standard SFT model might easily misinterpret ambiguous visual cues, leading to suboptimal or overly cautious decisions, such as unnecessary deceleration, which disrupts traffic flow and fundamentally compromises safety.

To mitigate these inherent reasoning deficiencies, recent state-of-the-art methodologies have increasingly turned to augmenting VLMs with Chain-of-Thought (CoT)~\cite{wei2022chain} processes and explicit external tool calls. Frameworks such as AgentThink~\cite{agentthink} and OmniDrive-R1~\cite{omnidriver1} attempt to actively ground visual elements and verify intermediate logic by interleaving multi-modal reasoning with dynamic Application Programming Interface (API) calls, such as open-vocabulary detectors or depth estimators. However, as depicted in~\Cref{fig:intro}(b), this tool-augmented paradigm fundamentally compromises the robustness of the end-to-end architecture. It introduces a fragile dependency pipeline where a single point of failure or a noisy output from an external module can cascade, causing the entire decision-making process to collapse or produce hazardous directives. Furthermore, generating hundreds of explicit reasoning tokens imposes prohibitive computational overhead, rendering these prolonged System-2 approaches impractical for real-time vehicular deployment.

To overcome the fragility of tool-augmented pipelines, we introduce a new Critique-Driven Multi-Turn Reinforcement Learning (RL) paradigm in the second stage of our CritiqueDriveVLM framework. Instead of relying on brittle external perception APIs, we construct a frozen Multi-Dimensional Verifier that strictly evaluates the model's reasoning trajectories across perception, logic, and safety dimensions. As illustrated in~\Cref{fig:intro}(c), this verifier acts as a logical guide during inference, providing targeted natural language critiques that allow the Teacher model to engage in a multi-turn interactive loop, iteratively refining its initial thoughts and correcting hallucinations. Concurrently, during the Group Relative Policy Optimization (GRPO)~\cite{shao2024deepseekmath} training phase, the verifier provides fine-grained scalar rewards, while a step-decay multi-turn penalty explicitly forces the policy to internalize the logical deduction process. This cultivates a highly reliable Teacher model that achieves profound System-2 reasoning and robust decision-making through verifier-guided refinement.

While our Stage-2 Teacher achieves state-of-the-art accuracy, the heavy computational burden of explicit CoT reasoning limits its real-time applicability. Conventional knowledge distillation~\cite{distillation} typically forces a lightweight student either to mimic the teacher's verbose CoT tokens, failing to resolve the latency bottleneck, or to directly replicate the final answer, severely truncating the required logical depth. To bridge this gap, we introduce a novel Latent Thought Distillation strategy in the third stage. Unlike tool-augmented methods that remain heavily bottlenecked by explicit API calls latency and fragile parsing pipelines, our proposed distillation operates entirely within the model's internal latent space. Rather than imitating surface-level text, we directly align the Student's latent representations with the Teacher's fully converged hidden states at the end of its reasoning trajectory, specifically at the final \texttt{</think>} token. As depicted in~\Cref{fig:intro}(d), this novel objective effectively compresses the complex reasoning capabilities of the System-2 Teacher into a fast, CoT-free System-1 Student. The distilled Student completely bypasses both the external verifier and the intermediate text generation, predicting the safe driving action with ultra-low latency, thereby achieving an optimal balance between deep reasoning rigor and low-latency inference efficiency.

The main contributions of this work can be summarized as follows:
\begin{enumerate}
    \item We propose \textbf{CritiqueDriveVLM}, a novel three-stage framework specifically designed to bridge the gap between deep System-2 logic and low-latency System-1 execution for autonomous driving VLMs.
    \item We introduce \textbf{Critique-Driven Multi-Turn RL} equipped with a multi-dimensional verifier and step-decay penalty, successfully suppressing visual hallucinations without relying on fragile external tools.
    \item We develop \textbf{Latent Thought Distillation} to align the student's hidden representations with the teacher's fully converged reasoning states, enabling highly accurate CoT-free inference.
    \item Extensive experiments on the \textbf{DriveLMM-o1}~\cite{drivelmmo1} benchmark demonstrate the superiority of our approach. Our tool-free Teacher model achieves state-of-the-art reasoning accuracy, while our distilled Student model drastically slashes inference latency to low-latency levels, paving a practical pathway for robust autonomous driving.
\end{enumerate}

\section{Related Work}

\subsection{VLMs in Autonomous Driving}
The integration of VLMs has shifted autonomous driving from modular~\cite{fan2018baidu,ettinger2021large} pipelines to end-to-end holistic reasoning~\cite{jiang2023vad,hu2023planning,wang2023drivemlm,t2sg,selfdrivingattn,ttabn}. Pioneering works such as DriveVLM~\cite{DriveVLM,nie2024reason2drive} demonstrate the potential of VLMs in complex scene understanding and hierarchical planning. To systematically evaluate and advance these capabilities, benchmarks such as DriveLMM-o1~\cite{drivelmmo1,sima2024drivelm} have been introduced, providing specialized datasets that emphasize step-wise visual reasoning and deep logical inference in complex dynamic scenarios. To mitigate reasoning hallucinations, recent state-of-the-art approaches rely on tool-augmented architectures and prolonged CoT. For instance, AgentThink~\cite{agentthink} integrates CoT with dynamic tool calls, while OmniDrive-R1~\cite{omnidriver1} employs an interleaved Multi-modal CoT (iMCoT) mechanism for fine-grained active visual grounding. However, these methods incur exorbitant token consumption and latency bottlenecks~\cite{safedriverag}. In contrast, our proposed CritiqueDriveVLM eliminates dependencies on external APIs and explicit CoT generation, achieving robust low-latency inference.

\subsection{Reinforcement Learning with Verifiable Rewards}
Reinforcement Learning with Verifiable Rewards (RLVR) has emerged as a powerful paradigm to unlock the reasoning potential of Large Language Models~\cite{ouyang2022training,lightman2024lets}. Pioneering works such as DeepSeekMath~\cite{shao2024deepseekmath} and DeepSeekMathV2~\cite{deepseekmathv2} demonstrate that RL, particularly through algorithms such as GRPO, can significantly boost logical performance in structured domains by optimizing against objective, rule-based rewards. DeepSeek-R1~\cite{deepseekr1} further showcases how large-scale RLVR elicits emergent self-correction and sophisticated reasoning trajectories. Recently, AlphaDrive~\cite{alphadrive} introduces a two-stage GRPO-based RL framework tailored for autonomous driving, demonstrating that RL can effectively elicit emergent multimodal planning capabilities beyond standard SFT~\cite{poutine,autovla,pag,pigeon}. However, applying RLVR to complex, open-ended driving scenarios often suffers from unreliable or sparse reward signals, inevitably leading to severe hallucinations~\cite{li2023evaluating} during the CoT process. To address this, our framework introduces a multi-dimensional verifier to provide granular feedback during RL, effectively suppressing CoT hallucinations and establishing a highly reliable Teacher model capable of critique-driven refinement for subsequent distillation.

\subsection{Knowledge Distillation}
Knowledge Distillation~\cite{distillation,romero2014fitnets,dcsam,rdcl,actionform,aqa,balanced3dpose,crossmodalretrieval,fewshotensemble,sportscaption} is fundamentally designed to transfer capabilities from a cumbersome teacher model to a lightweight student. Early alignment methods, such as Self-Instruct~\cite{selfinstruct}, primarily relied on bootstrapping instruction-following data to mimic output distributions. Subsequent advancements, including Distilling Step-by-Step~\cite{distillingstep} and MiniLLM~\cite{minillm}, explicitly distilled CoT rationales to enhance generation quality in smaller models. Recently, researchers have explored internalizing slow System 2 reasoning into fast System 1 intuition~\cite{yu2024distilling, kahneman, coconut, codi, icotkd}, while the distillation phase of DeepSeek-R1~\cite{deepseekr1} demonstrates that smaller models can successfully inherit complex reasoning capabilities from large RL-trained models. However, existing distillation paradigms typically force the student model to either mimic the teacher's explicit CoT tokens or simply replicate the final response at the token level. In autonomous driving, generating even a shortened CoT trajectory incurs unacceptable latency, while standard token-level imitation fails to transfer the intricate spatial and logical reasoning deeply embedded in the teacher's cognitive process. To address this, we propose a new Latent Thought Distillation approach that directly aligns the student's hidden representations with the teacher's fully converged reasoning states, achieving an optimal balance between reasoning depth and inference efficiency.

\section{Method}
\begin{figure}[tb]
  \centering
  \includegraphics[width=1.0\textwidth]{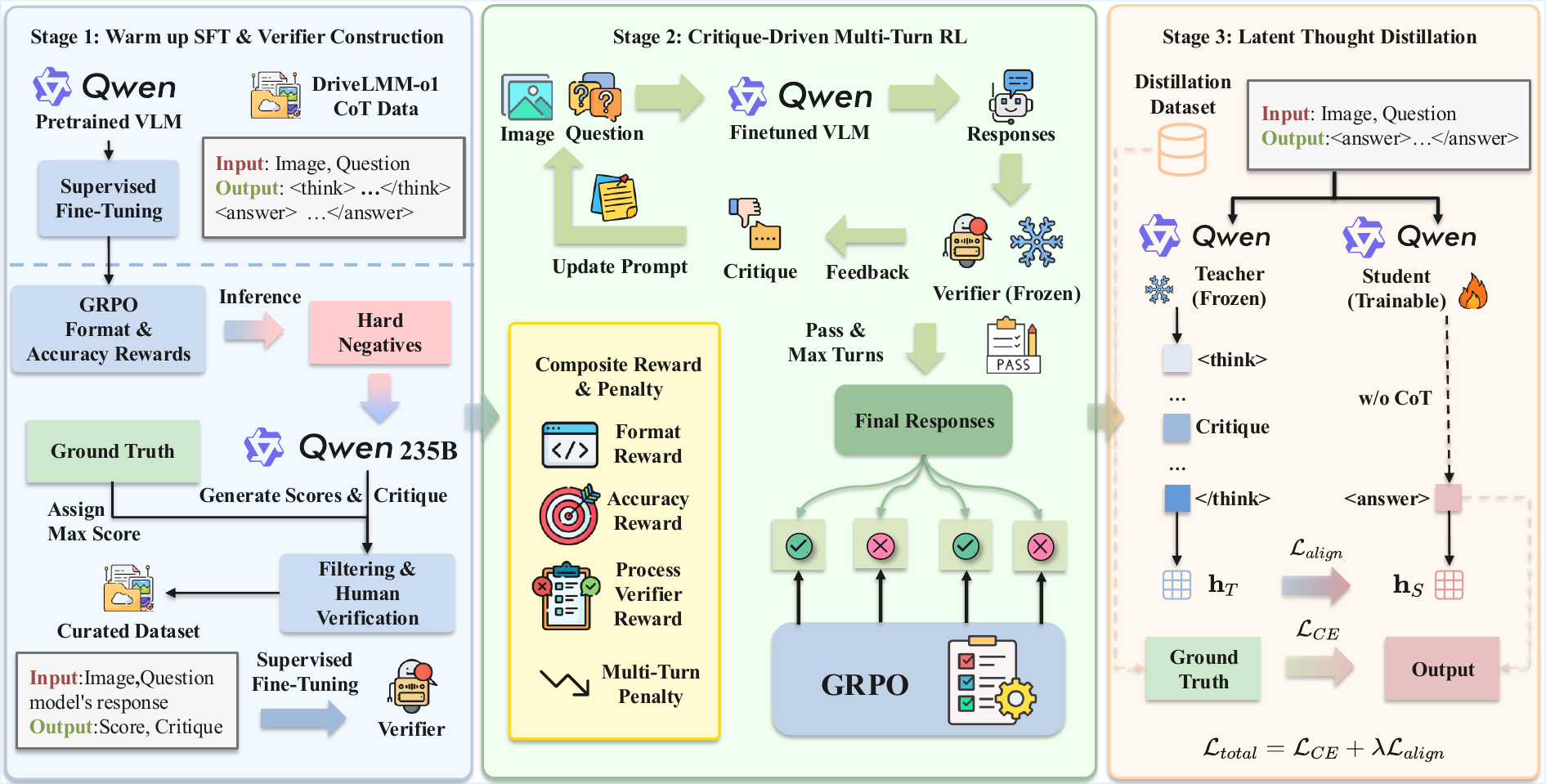}
\caption{Overall pipeline of the proposed CritiqueDriveVLM framework.
(i) Stage 1: Warm-up SFT and Verifier Construction establishing a structural reasoning format and training a multi-dimensional verifier;
(ii) Stage 2: Critique-Driven Multi-Turn RL cultivating a highly reliable Teacher via verifier feedback; and 
(iii) Stage 3: Latent Thought Distillation internalizing the Teacher's latent representations into a CoT-free, low-latency Student model.}
  \label{fig:fig2}
\end{figure}

In this section, we describe the proposed CritiqueDriveVLM framework, designed to tackle two critical bottlenecks in autonomous driving VLMs: the inherent unreliability of reasoning in complex scenarios, and the unacceptable inference latency of explicit CoT generation. As illustrated in~\Cref{fig:fig2}, our framework is implemented through a three-stage training pipeline:
\textbf{Stage 1} (Warm-up SFT and Verifier Construction) establishes the foundational reasoning format and builds a multi-dimensional verifier; 
\textbf{Stage 2} (Critique-Driven Multi-Turn RL) leverages verifier feedback and a multi-turn penalty to cultivate a highly reliable Teacher model capable of verifier-guided refinement; Finally, \textbf{Stage 3} (Latent Thought Distillation) transfers the Teacher's deep logical representations into a Student model, eliminating the latency of explicit CoT generation.

\subsection{Warm-up SFT and Verifier Construction}
Before initiating large-scale RL, it is essential to equip the base policy with a standardized CoT paradigm and construct an independent multi-dimensional verifier.

\subsubsection{Warm-up SFT.}
We perform a warm-up SFT on the base VLM using a subset of high-quality CoT data from the DriveLMM-o1 dataset~\cite{drivelmmo1}. The primary objective here is not to maximize the model's reasoning capabilities, but to strictly enforce a structural output format. Given an input image $V$ and a user instruction $x$, the model is trained to consistently generate a complete response $y$ formatted as $y = \texttt{<think>}\ r\ \texttt{</think>}\ \texttt{<answer>}\ a\ \texttt{</answer>}$, where $r$ denotes the intermediate reasoning process and $a$ represents the final answer. This format initialization acts as a critical anchor, preventing mode collapse and format degradation when the model explores the expansive action space during the subsequent RL phase.

\subsubsection{Multi-Dimensional Verifier Construction.}
The core component of our multi-turn framework is an independent, frozen multi-dimensional verifier $\mathcal{V}$, responsible for evaluating the model's complete generated response $y$. To train this verifier (initialized from a separate base model), we design a rigorous data generation pipeline. Positive samples are directly derived from ground truth annotations, which are inherently assigned maximum scalar scores across all dimensions with an empty natural language critique. To mine hard negatives, we deploy a baseline GRPO~\cite{shao2024deepseekmath} model, trained solely with formatting and accuracy rewards, to perform inference on the training set. This yields a wealth of erroneous outputs exhibiting visual hallucinations or logical contradictions. These flawed responses are subsequently evaluated by a substantially more capable model (Qwen3-VL-235B~\cite{bai2025qwen3}), which assigns discrete multi-dimensional scores as follows:
\begin{equation}
    R_{\text{verif}} = \{s_{\text{per}}, s_{\text{log}}, s_{\text{saf}}\}, \quad s \in \{0, 0.5, 1.0\},
\end{equation}
covering perception, logic, and safety. Concurrently, the Qwen3-VL-235B model generates targeted text-based critiques $c$ addressing the specific errors. Following a stringent data filtering and human verification process, this curated dataset is used to train the verifier. During the RL stage, the function $\mathcal{V}(V, x, y) \rightarrow (R_{\text{verif}}, c)$ provides the policy model with granular scalar feedback and, if the response fails to meet requirements, it also generates a text-based critique prompt to guide the subsequent reasoning turn.

\subsection{Critique-Driven Multi-Turn RL}
To train a highly reliable Teacher model $\pi_\theta$ capable of profound logical reasoning and verifier-guided refinement, we introduce a multi-turn RL paradigm. Unlike traditional RL settings that rely solely on binary task-completion rewards, our framework leverages the frozen multi-dimensional verifier $\mathcal{V}$ (constructed in Stage 1) to provide fine-grained textual and scalar feedback. The definitions of the composite reward components are detailed in~\Cref{tab:reward_definition}.

\begin{table}[tb]
    \caption{Definitions of the composite reward components. By isolating the reasoning process (\texttt{<think>}) and final decision (\texttt{<answer>}), our Multi-Turn RLVR provides fine-grained feedback to mitigate conservative bias and visual hallucinations.}
  \label{tab:reward_definition}
  \centering
  \footnotesize 
  \begin{tabularx}{\linewidth}{@{} l X @{}}
    \toprule
    \textbf{Reward Type} & \textbf{Description} \\
    \midrule
    Format Reward ($R_{\text{fmt}}$) & \textbf{Format Compliance:} Ensures the model's output strictly adheres to the predefined XML-style structure (\ie, isolating reasoning within \texttt{<think>} tags and decisions within \texttt{<answer>} tags). Promotes reliable trajectory parsing. \\ 
    \midrule
    Accuracy Reward ($R_{\text{acc}}$) & \textbf{Final Answer Alignment:} Verifies the extracted final action directive (parsed from the \texttt{<answer>} block) against the ground truth. Promotes task-level decision correctness. \\ 
    \midrule
    Process Verifier Reward ($R_{\text{verif}}$) & Evaluates the intermediate CoT (parsed from the \texttt{<think>} block) through multi-dimensional scalar feedback. Sub-rewards for: \par
    \textbf{(i) Perception Score ($s_{\text{per}}$):} Verifies visual entity grounding accuracy and penalizes visual hallucinations. \par
    \textbf{(ii) Logic Score ($s_{\text{log}}$):} Evaluates the rigor of causal reasoning and penalizes internal logical contradictions. \par
    \textbf{(iii) Safety Score ($s_{\text{saf}}$):} Assesses reasoning safety to mitigate conservative bias. \\ 
    \midrule
    Multi-Turn Penalty ($\mathcal{P}_{\text{mt}}$) & \textbf{Efficiency Constraint:} A step-decay penalty applied based on the number of critique attempt turns. Forces the model to internalize reasoning and promotes first-attempt correctness. \\
    \bottomrule
  \end{tabularx}
\end{table}

\subsubsection{Multi-Turn Interaction.} 
We formulate the reasoning process as a sequential interactive loop with a maximum allowed turn limit $K$. At the initial turn $t=1$, given an input image $V$ and a prompt instruction $x_1 = x$, the policy model generates a complete response $y_1$. The verifier $\mathcal{V}$ then evaluates $y_1$ to output a score set $R_{\text{verif}}$ and a text critique $c_1$. If $R_{\text{verif}}$ achieves a perfect score across all dimensions, or the interaction reaches the maximum limit ($t = K$), the loop terminates at turn $T$. Otherwise, the verifier outputs a critique $c_t$, which is appended to the dialogue history to construct a new user prompt: $x_{t+1} = \text{Concat}(x_t, y_t, c_t)$. The model then generates a refined response $y_{t+1}$ based on the critique history~\cite{shinn2023reflexion,madaan2023self}.

\subsubsection{Composite Reward and Efficiency Penalty.} 
The reward is calculated based on the terminal response $y_T$ at the concluding turn $T$. As outlined in~\Cref{tab:reward_definition}, the base reward $R_{\text{base}}$ aggregates formatting, accuracy, and process-aware evaluations as follows:
\begin{equation}
    R_{\text{base}}(y_T) = R_{\text{fmt}}(y_T) + R_{\text{acc}}(y_T) + \alpha (s_{\text{per}} + s_{\text{log}} + s_{\text{saf}}),
\end{equation}
where $\alpha$ is a weighting coefficient for the verifier scores. 

While multi-turn interaction significantly enhances reasoning accuracy, autonomous driving systems demand extremely low latency. To prevent the model from becoming overly reliant on external critiques, we introduce a step-decay multi-turn penalty $\mathcal{P}_{\text{mt}}(T)$. This penalty explicitly forces the model to internalize its reasoning process and promotes first-attempt correctness. The final terminal reward is formulated as:
\begin{equation}
    R_{\text{final}} = R_{\text{base}}(y_T) - \mathcal{P}_{\text{mt}}(T).
\end{equation}
In practical implementation, to optimally balance exploration efficiency and computational overhead, we set the maximum turn limit to $K=2$. Thus, the model is penalized with a constant decay value if it requires a second turn to correct its errors.

\subsubsection{GRPO Optimization Objective.}
To avoid the massive memory overhead associated with maintaining an equivalent-sized Critic network in the standard Proximal Policy Optimization (PPO) algorithm~\cite{schulman2017proximal}, we employ GRPO in this work. For a given instruction $x$ drawn from the training distribution $P(x)$, we sample a group of $G$ terminal responses, denoted as $\{y_1, y_2, \dots, y_G\}$, from the old policy $\pi_{\theta_{\text{old}}}$. The policy is optimized by maximizing the following GRPO objective:
\begin{equation}
    \mathcal{J}_{\text{GRPO}}(\theta) = \mathbb{E}_{x \sim P(x), \{y_i\}_{i=1}^G \sim \pi_{\theta_{\text{old}}}} \left[ \frac{1}{G} \sum_{i=1}^G \mathcal{L}_i - \beta \mathbb{D}_{\text{KL}}(\pi_\theta \| \pi_{\text{ref}}) \right],
\end{equation}
where the group-wise clipped loss $\mathcal{L}_i$ is defined as follows:
\begin{equation}
    \mathcal{L}_i = \min \left( \rho_i A_i, \operatorname{clip}(\rho_i, 1-\epsilon, 1+\epsilon) A_i \right),
\end{equation}
where the importance sampling ratio $\rho_i$ and the normalized advantage $A_i$ are given by:
\begin{align}
    \rho_i &= \frac{\pi_\theta(y_i|x)}{\pi_{\theta_{\text{old}}}(y_i|x)}, \\
    A_i &= \frac{R_{\text{final}}^{(i)} - \operatorname{mean}(R_{\text{final}})}{\operatorname{std}(R_{\text{final}})}.
\end{align}
The hyperparameter $\epsilon$ clips this ratio to ensure training stability. Additionally, $\beta$ controls the strength of the Kullback-Leibler (KL) divergence penalty $\mathbb{D}_{\text{KL}}$, which restricts the policy $\pi_\theta$ from deviating excessively from the reference model $\pi_{\text{ref}}$ established during the warm-up SFT stage.

\subsection{Latent Thought Distillation}
To resolve the prohibitive latency of explicit CoT generation, we propose Latent Thought Distillation, internalizing the Teacher's slow System-2 reasoning into the Student's fast System-1 intuition without outputting CoT text, inspired by~\cite{kahneman,yu2024distilling}.

We initialize the Student model directly from the original base VLM. Furthermore, we construct the distillation dataset by strictly formatting the DriveLMM-o1 training data into pure prompt-answer pairs, entirely removing any explicit CoT text. This setup forces the Student to rely solely on its internal latent representations to maintain logical depth.

Our strategy aligns the hidden representation of the Student with the final reasoning state of the Teacher. Let $\pi_T$ and $\pi_S$ be the frozen Teacher and trainable Student, respectively. For a given input, $\pi_T$ generates its multi-turn reasoning trajectory. We extract the deep hidden state at the last \texttt{</think>} token of the trajectory, denoted as $\mathbf{h}_{T}^{\text{think}} \in \mathbb{R}^{d}$. Simultaneously, $\pi_S$ skips the intermediate reasoning steps to directly predict the final output. We extract its corresponding hidden state at the \texttt{<answer>} token, denoted as $\mathbf{h}_{S}^{\text{answer}} \in \mathbb{R}^{d}$.

To ensure the Student's internal logic mimics the Teacher's thought process, we employ Cosine Similarity as the alignment objective:
\begin{equation}
    \mathcal{L}_{\text{align}} = 1 - \frac{\mathbf{h}_{S}^{\text{answer}} \cdot \mathbf{h}_{T}^{\text{think}}}{\|\mathbf{h}_{S}^{\text{answer}}\|_2 \|\mathbf{h}_{T}^{\text{think}}\|_2}.
\end{equation}
This objective strictly focuses on aligning the semantic direction of the reasoning logic in high-dimensional space. The Student model is subsequently optimized through a joint loss function:
\begin{equation}
    \mathcal{L}_{\text{total}} = \mathcal{L}_{\text{CE}} + \lambda \mathcal{L}_{\text{align}},
\end{equation}
where $\mathcal{L}_{\text{CE}}$ is the standard auto-regressive cross-entropy loss for supervised answer prediction, and $\lambda$ is the balancing coefficient. Through this unified objective, the Student inherits the Teacher's logical rigor within its latent space, achieving efficient and accurate prediction.

\section{Experiments}
\label{sec:experiments}

In this section, we comprehensively evaluate the effectiveness of the proposed CritiqueDriveVLM framework in complex autonomous driving scenarios. The remainder of this section is organized as follows: We first detail the experimental setup in~\Cref{subsec:setup}. Then, we present the main quantitative results on the DriveLMM-o1 benchmark~\cite{drivelmmo1}, highlighting the state-of-the-art reasoning capabilities of our Stage-2 Teacher model in~\Cref{subsec:main_results}. Next, we analyze the inference efficiency and the impact of Latent Thought Distillation under CoT-free constraints in~\Cref{subsec:efficiency}. We further conduct ablation studies to validate the critique-driven multi-turn reinforcement learning mechanisms in~\Cref{subsec:ablation}. Finally, we provide a qualitative analysis of our model's decision-making process in~\Cref{subsec:qualitative}.

\subsection{Experimental Setup}
\label{subsec:setup}

\subsubsection{Datasets.}
To ensure a fair comparison and strictly validate the effectiveness of our framework, we exclusively utilize the DriveLMM-o1~\cite{drivelmmo1} benchmark, which is built upon the foundational nuScenes~\cite{Caesar_2020_CVPR} dataset. Specifically, we employ its training split comprising over 18,000 VQA pairs, where each sample provides step-by-step reasoning for complex driving scenarios. This unified dataset is consistently applied across all three stages of our framework: Warm-up SFT, Critique-Driven RL, and Latent Thought Distillation.

\subsubsection{Evaluation Metrics.}
We adopt the official evaluation protocol of the DriveLMM-o1 benchmark, reporting the Overall Reasoning score and Multiple Choice Quality (MCQ) as primary metrics. Additionally, we evaluate fine-grained task-specific driving and scene detail metrics (\eg, Risk Assessment, Traffic Rule Adherence, Scene Awareness, Relevance, and Missing Details). Further metric calculation details are provided in Appendix~A.

\subsubsection{Implementation Details.}
We employ Qwen3-VL-8B~\cite{bai2025qwen3} as our base model for both the Teacher and the Student, keeping the vision encoder frozen across all training phases. During Stage 1, the Warm-up SFT is conducted using Low-Rank Adaptation (LoRA)~\cite{hu2022lora} with a learning rate of $1 \times 10^{-4}$. For Stage 2, the GRPO~\cite{shao2024deepseekmath} fine-tuning utilizes a learning rate of $2 \times 10^{-6}$ with 4 rollouts per question ($G=4$), and the maximum multi-turn interaction limit is set to $K=2$. In Stage 3, the Latent Thought Distillation is also optimized using LoRA with a learning rate of $2 \times 10^{-5}$, and the balancing coefficient for the alignment loss is set to $\lambda = 0.5$. Across all three training phases, the models are optimized for 2 epochs with a global batch size of 128. All experiments are conducted on a single server equipped with 8 NVIDIA A100 GPUs, and inference time is tested on a single A100 GPU. Additional implementation details and hyperparameter settings are elaborated in Appendix~B.

\begin{table}[tb]
  \caption{Main quantitative comparison on the DriveLMM-o1 benchmark. \textbf{Bold} indicates best. \underline{Underline} indicates second best. Our Stage-2 Teacher achieves state-of-the-art results without external tool-use.}
  \label{tab:main}
  \centering
  \footnotesize
  \setlength{\tabcolsep}{2.8pt} 
  
  \begin{tabular}{@{} l ccc cc cc @{}} 
    \toprule
    \multirow{2}{*}{\makecell[l]{Vision Language \\ Models}} & \multicolumn{3}{c}{Driving Metrics (\%) $\uparrow$} & \multicolumn{2}{c}{Scene Detail (\%) $\uparrow$} & \multicolumn{2}{c}{Overall (\%) $\uparrow$} \\
    \cmidrule(r){2-4} \cmidrule(lr){5-6} \cmidrule(l){7-8}
    & \makecell{Risk\\Assess.} & \makecell{Rule\\Adh.} & \makecell{Scene\\Aware.} & \makecell{Relevance} & \makecell{Missing} & \makecell{Reason.} & MCQ \\
    \midrule
    GPT-4o~\cite{islam2025gpt}             & 71.32 & 80.72 & 72.96 & 76.65 & 71.43 & 72.52 & 57.84 \\
    Ovis1.5-Gemma2-9B~\cite{lu2024ovis}  & 51.34 & 66.36 & 54.74 & 55.72 & 55.74 & 55.62 & 48.85 \\
    Mulberry-7B~\cite{yao2024mulberry}      & 51.89 & 63.66 & 56.68 & 57.27 & 57.45 & 57.65 & 52.86 \\
    LLaVA-CoT~\cite{xu2025llava}          & 57.62 & 69.01 & 60.84 & 62.72 & 60.67 & 61.41 & 49.27 \\
    LlamaV-O1~\cite{thawakar2025llamav}         & 60.20 & 73.52 & 62.67 & 64.66 & 63.41 & 63.13 & 50.02 \\
    InternVL2.5-8B~\cite{chen2025internvl}     & 69.02 & 78.43 & 71.52 & 75.80 & 70.54 & 71.62 & 54.87 \\
    Qwen2.5-VL-7B~\cite{bai2025qwen25vl}       & 46.44 & 60.45 & 51.02 & 50.15 & 52.19 & 51.77 & 37.81 \\
    Qwen2.5-VL-72B~\cite{bai2025qwen25vl}      & 64.40 & 72.81 & 60.29 & 65.13 & 62.81 & 65.73 & 61.27 \\
    Qwen3-VL-8B~\cite{bai2025qwen3}         & 79.50 & 84.32 & 80.41 & 84.43 & 75.32 & 77.76 & 55.54 \\
    DriveLMM-o1~\cite{drivelmmo1}        & 73.01 & 81.56 & 75.39 & 79.42 & 74.49 & 75.24 & 62.36 \\
    AgentThink~\cite{agentthink} & 80.51 & 84.98 & 82.11 & \underline{84.99} & \textbf{79.56} & 79.68 & 71.35 \\
    OmniDrive-R1~\cite{omnidriver1} & \textbf{82.31} & \underline{85.42} & \textbf{83.75} & 82.58 & 78.26 & \underline{80.35} & \underline{73.62} \\
    \rowcolor[gray]{0.92} \textbf{Ours-Teacher(Stage 2)} & \underline{81.63} & \textbf{86.21} & \underline{82.27} & \textbf{85.46} & \underline{79.17} & \textbf{80.48} & \textbf{76.54} \\
    \bottomrule
  \end{tabular}
\end{table}

\subsection{Main Results}
\label{subsec:main_results}

We evaluate the comprehensive driving and reasoning capabilities of our framework against a wide range of state-of-the-art VLMs on the DriveLMM-o1 benchmark. As shown in~\Cref{tab:main}, the comparative baselines include prominent general-purpose models (\eg, GPT-4o~\cite{islam2025gpt}, InternVL2.5-8B~\cite{chen2025internvl}, and the Qwen-VL series~\cite{bai2025qwen25vl,bai2025qwen3}) as well as domain-specific autonomous driving models (\eg, DriveLMM-o1~\cite{drivelmmo1}, AgentThink~\cite{agentthink}, and OmniDrive-R1~\cite{omnidriver1}).

Our Stage-2 Teacher establishes a new state-of-the-art across multiple critical metrics. Notably, it achieves the highest Overall Reasoning score of 80.48\% and an MCQ of 76.54\%, significantly outperforming strong driving-specific counterparts like OmniDrive-R1 (73.62\%) and AgentThink (71.35\%). Unlike baselines relying on complex external APIs or interleaved multi-modal CoT, our Teacher achieves superior accuracy entirely through internalized logic cultivated during the critique-driven RL phase, avoiding explicit tool-use overhead.

Delving into the fine-grained metrics, our model demonstrates a significant advantage in Traffic Rule Adherence (86.21\%) and Relevance (85.46\%). While OmniDrive-R1 marginally leads in Risk Assessment (82.31\%) and Scene Awareness and Object Understanding (83.75\%), and AgentThink slightly edges out in the Missing Details metric (79.56\%), our Teacher model maintains highly competitive and robust scores in these areas (81.63\%, 82.27\%, and 79.17\%, respectively). This comprehensive and well-rounded performance profile proves that providing granular, multi-dimensional verifier feedback during RL effectively suppresses visual hallucinations and conservative biases, resulting in highly reliable end-to-end driving decisions.

\begin{table}[tb]
  \caption{Comparison of inference efficiency and ablation on Latent Thought Distillation (Stage 3). We evaluate the trade-off between reasoning accuracy and latency. The ablation of the alignment objective ($\mathcal{L}_\text{align}$) demonstrates that our Student model preserves reasoning depth while achieving low-latency execution under CoT-free constraints.}
  \label{tab:efficiency_ablation}
  \centering
  \footnotesize 
  \setlength{\tabcolsep}{2.5pt} 
  \begin{tabular}{@{} l | ccc | c c c @{}}
    \toprule
    Models & CoT & Tool & $\mathcal{L}_\text{align}$ &  Avg. Tokens $\downarrow$ & \makecell{Avg. Time \\ (ms) $\downarrow$} & MCQ (\%) $\uparrow$ \\
    \midrule
    \multicolumn{7}{c}{\textbf{System-2 Models (with CoT)}} \\
    \midrule
    Qwen3-VL-8B                    & \checkmark & $\times$   & $\times$ & 390.95 & 4265 & 55.54      \\ 
    Qwen3-VL-8B (SFT)              & \checkmark & $\times$   & $\times$ & 147.88 & 1675 & 62.86      \\
    DriveLMM-o1                    & \checkmark & $\times$   & $\times$ & 150.52 & 1968 & 62.36      \\
    AgentThink                     & \checkmark & \checkmark & $\times$ & 515.01 & 5416 & 71.35      \\
     \textbf{Ours-Teacher(Stage 2)}         & \checkmark & $\times$   & $\times$ & 223.32 & 3482 & \textbf{76.54} \\
    \midrule
    \multicolumn{7}{c}{\textbf{System-1 Models (CoT-Free)}} \\
    \midrule
    Qwen3-VL-8B                    & $\times$   & $\times$   & $\times$ & 37.23 & 461  & 54.96      \\
    Qwen3-VL-8B (SFT)              & $\times$   & $\times$   & $\times$ & 23.65 & 343  & 61.73      \\ 
    DriveLMM-o1                    & $\times$   & $\times$   & $\times$ & 30.96 & 425  & 60.14      \\
     \textbf{Ours-Student(Stage 3)}         & $\times$   & $\times$   & \checkmark & 28.83 & 416  & \textbf{68.59} \\
    \bottomrule
  \end{tabular}
\end{table}

\subsection{Efficiency and Latent Thought Distillation}
\label{subsec:efficiency}

Real-time autonomous driving systems impose stringent constraints on inference latency. While explicit CoT generation and tool-use mechanisms significantly enhance reasoning accuracy, they inherently incur unacceptable computational overhead. As shown in the System-2 section of~\Cref{tab:efficiency_ablation}, traditional reasoning models suffer from extreme verbosity. AgentThink incurs a staggering latency of 5416 ms due to explicit tool invocations. Even our highly capable Stage-2 Teacher requires 3482 ms to articulate its internal reasoning process. To eliminate this latency bottleneck, we must evaluate models under a strict CoT-free constraint.

However, when standard baselines are forced to predict final answers directly without CoT, their performance drops precipitously. As shown in the System-1 section of~\Cref{tab:efficiency_ablation}, the Qwen3-VL-8B (SFT) model without distillation severely truncates the logical depth, achieving an MCQ of only 61.73\%. To bridge this capability gap, our Latent Thought Distillation integrates an alignment objective ($\mathcal{L}_\text{align}$). By explicitly aligning the Student's latent representations with the Teacher's fully converged reasoning state, our Stage-3 Student achieves a superior MCQ of 68.59\% using an average of only 28.83 tokens. This drastically slashes the absolute inference time to 416 ms, an 88\% reduction compared to the Stage-2 Teacher (3482 ms), demonstrating that our proposed distillation method successfully internalizes deep System-2 reasoning capabilities into the fast System-1 latent space, achieving an optimal balance between reasoning accuracy and low-latency execution.

\subsection{Ablation on Critique-Driven Multi-Turn RL}
\label{subsec:ablation}

To validate the contribution of each core mechanism in Critique-Driven Multi-Turn RL (Stage 2), we ablate our reward design in~\Cref{tab:rl_ablation}. Starting from the Warm-up SFT baseline (62.86\% MCQ), applying standard single-turn GRPO with only Accuracy Rewards yields a marginal improvement (64.49\% MCQ). The addition of Multi-dimensional Verifier Rewards further boosts the MCQ to 67.46\% while steadily improving fine-grained metrics such as Risk Assessment and Traffic Rule Adherence, indicating that providing granular scalar feedback effectively suppresses visual hallucinations. Crucially, introducing the Multi-Turn Interaction combined with a step-decay penalty significantly elevates the overall MCQ to 76.54\%. This substantial improvement confirms our hypothesis: transitioning from a single-turn setup to a multi-turn loop allows the model to iteratively refine its reasoning based on verifier critiques, while the multi-turn penalty prevents over-reliance on these external cues, forcing the policy to internalize the logical deduction process for correctness.

\begin{table}[tb]
  \caption{Ablation study on Critique-Driven Multi-Turn RL (Stage 2). We progressively evaluate the impact of Accuracy Rewards (Acc), Multi-dimensional Verifier Rewards (Ver), and Multi-Turn Interaction \& Penalty (MT).}
  \label{tab:rl_ablation}
  \centering
  \scriptsize 
  \setlength{\tabcolsep}{3.5pt} 
  \begin{tabular}{@{} l | c | ccc | ccc | cc | cc @{}}
    \toprule
    \multirow{3}{*}{Model Variant} & \multirow{3}{*}{SFT} & \multicolumn{3}{c|}{Reward Setting} & \multicolumn{3}{c|}{Driving Metrics (\%) $\uparrow$} & \multicolumn{2}{c|}{Detail (\%) $\uparrow$} & \multicolumn{2}{c}{Overall (\%) $\uparrow$} \\
    & & Acc & Ver & MT & \makecell{Risk\\Assess.} & \makecell{Rule\\Adh.} & \makecell{Scene\\Aware.} & Relev. & Miss. & Reason. & MCQ \\
    \midrule
        Base Model             & $\times$   & $\times$   & $\times$   & $\times$   & 79.50 & 84.32 & 80.41 & 84.43 & 75.32 & 77.76 & 55.54 \\
        + SFT                  & \checkmark & $\times$   & $\times$   & $\times$   & 73.74 & 79.46 & 75.19 & 79.70 & 71.24 & 73.38 & 62.86 \\
        + GRPO                 & \checkmark & \checkmark & $\times$   & $\times$   & 74.14 & 80.12 & 75.59 & 80.21 & 71.63 & 73.85 & 64.49 \\
        + GRPO$^\dagger$       & \checkmark & \checkmark & \checkmark & $\times$   & 75.30 & 81.20 & 76.84 & 81.23 & 72.89 & 75.10 & 67.46 \\
        \textbf{Ours-Teacher}  & \checkmark & \checkmark & \checkmark & \checkmark & \textbf{81.63} & \textbf{86.21} & \textbf{82.27} & \textbf{85.46} & \textbf{79.17} & \textbf{80.48} & \textbf{76.54} \\
    \bottomrule
  \end{tabular}
\end{table}

\begin{figure}[tb]
  \centering
  \includegraphics[width=0.96\textwidth]{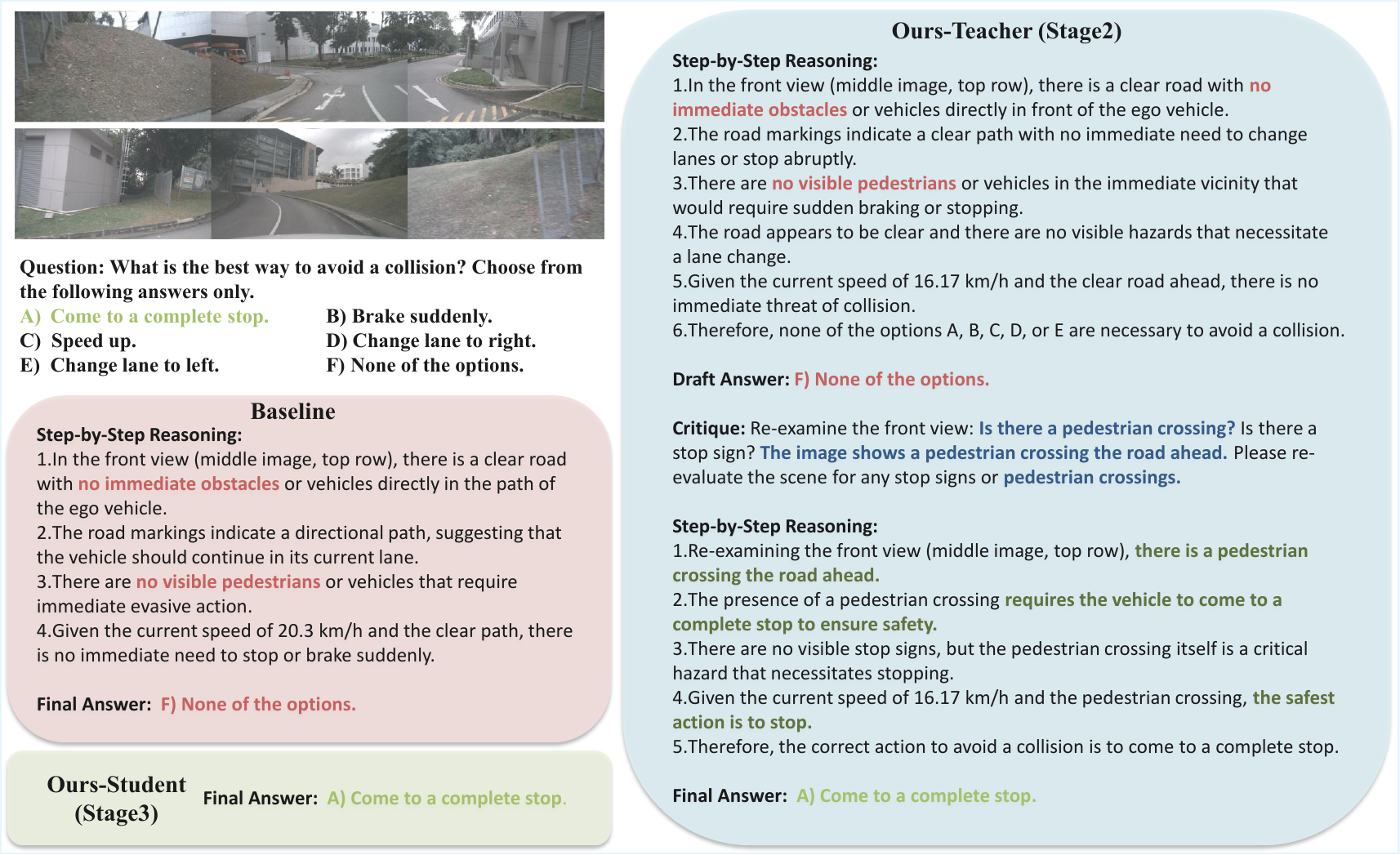}
  \caption{
  Qualitative comparison in a pedestrian crossing scenario. The \textbf{Baseline} hallucinates and misses the pedestrian. Our \textbf{Stage-2 Teacher} corrects this error through verifier-guided critiques, while our \textbf{Stage-3 Student} directly predicts the safe action without explicit CoT overhead.
  }
  \label{fig:example}
\end{figure}

\subsection{Qualitative Analysis}
\label{subsec:qualitative}      

To intuitively demonstrate the efficacy of our CritiqueDriveVLM framework, \Cref{fig:example} shows a comparative case study in a complex driving scenario involving a pedestrian crossing. As illustrated, the standard SFT Baseline suffers from severe visual hallucination; it explicitly reasons that there are ``no visible pedestrians'' and dangerously concludes that no evasive action is required (Option F). In contrast, our Stage-2 Teacher overcomes this failure through its critique-driven multi-turn reasoning mechanism. While its initial draft mirrors the Baseline's dangerous oversight, the external verifier injects targeted critique feedback, alerting the model to the overlooked pedestrian. This prompts the Teacher to successfully re-examine the visual evidence, correct its flawed trajectory, and output the safe action to ``Come to a complete stop'' (Option A). Crucially, our Stage-3 Student directly and instantly predicts the correct, safe action (Option A) without any explicit reasoning overhead. This confirms that our Latent Thought Distillation effectively compresses the Teacher's rigorous System-2 risk assessment capabilities directly into the Student's fast System-1 latent space.

\section{Conclusion}
\label{sec:conclusion}

In this paper, we presented \textbf{CritiqueDriveVLM}, a unified three-stage framework designed to resolve the reasoning fragility and inference latency bottlenecks in autonomous driving VLMs. By introducing Critique-Driven Multi-Turn RL, our Teacher model internalized logical deduction to mitigate visual hallucinations, achieving state-of-the-art accuracy without relying on fragile external tools. Furthermore, our Latent Thought Distillation successfully compressed these deep System-2 reasoning capabilities into a CoT-free System-1 Student, preserving decision accuracy while reducing inference latency to low-latency levels (416 ms). In the future, we plan to extend this framework to multi-camera video streams for temporal logic distillation.


\section*{Acknowledgements}
This work is partly supported by the Funds for the Beijing Natural Science Foundation under Grant L243027, the NSFC Project under Grant 62572072 and Hainan Provincial Natural Science Foundation.

%
%
\bibliographystyle{splncs04}
\bibliography{main}
\end{document}